\documentclass[a4paper, 10pt, conference]{ieeeconf}      

\IEEEoverridecommandlockouts                              

\usepackage{graphicx} 
\usepackage{mathptmx} 
\usepackage{booktabs}
\usepackage{multirow}
\usepackage{array}
\usepackage{microtype}
\usepackage{cite}
\usepackage{amsmath} 
\usepackage{amssymb}  

\newcommand{\method}{TLC-DiT}

\title{\LARGE \bf
TLC-DiT: Task-Aligned Local Visual Conditioning \\ for Robust Multitask Robot Manipulation
}

\author{
Xianbo Cai$^{1}$,
Hideyuki Ichiwara$^{1,2}$,
Zihang Wang$^{3}$,
Yijun Lu$^{4,*}$,
and Tetsuya Ogata$^{1,5}$%
\thanks{$^{1}$Xianbo Cai, Hideyuki Ichiwara, and Tetsuya Ogata are with the Department of Intermedia Art and Science, Waseda University, Tokyo, Japan. 
{\tt\footnotesize hourensou369@fuji.waseda.jp}}%
\thanks{$^{2}$Hideyuki Ichiwara is with SB Intuitions Corp., Tokyo, Japan.}%
\thanks{$^{3}$Zihang Wang is with the Department of Electronic and Physical Systems, Waseda University, Tokyo, Japan. 
{\tt\footnotesize wang.zihang@akane.waseda.jp}}%
\thanks{$^{4}$Yijun Lu is with the Department of Computer Science and Engineering, Waseda University, Tokyo, Japan. 
{\tt\footnotesize yijun@ruri.waseda.jp}}%
\thanks{$^{5}$Tetsuya Ogata is with the National Institute of Advanced Industrial Science and Technology (AIST), Tokyo, Japan. 
{\tt\footnotesize ogata@waseda.jp}}%
\thanks{$^{*}$Corresponding author: Yijun Lu.}%
}

\begin{document}

\maketitle
\thispagestyle{empty}
\pagestyle{empty}

\begin{abstract}
Language-conditioned robot policies have made clear progress in multitask manipulation, but task-relevant local visual evidence usually stays hidden inside a visual backbone or attention layers. This leaves the policy difficult to inspect and fragile under visual change—two symptoms of a missing explicit, task-aligned local visual channel. We present TLC-DiT, a plug-in extension of the Multitask Diffusion Transformer (DiT) policy that adds explicit task-guided local visual feature maps without changing the diffusion objective or the action-generation process. For each camera view, frozen DINOv2 patch features are modulated by the CLIP task embedding through FiLM and refined by a lightweight CoordConv CNN adapter into smooth spatial maps, which are concatenated with the original global image, language, joint-state, and timestep conditions. On LIBERO, TLC-DiT reaches a 93.5\% average success rate, compared with 86.5\% for Multitask DiT and 79.25\% for SmolVLA. On LIBERO-plus, the total success rate improves from 54.07\% to 57.24\%, with larger gains under camera, background, and sensor-noise changes. In real-world bimanual tasks, TLC-DiT raises Teabag Putting completion from 44\% to 89\% while maintaining comparable Match Box Opening performance. Feature-map visualizations confirm that the model attends to task-relevant regions across views and perturbations, providing a direct way to inspect the visual evidence.

\end{abstract}

\section{INTRODUCTION}

Robot policy learning has moved from single-task behavior cloning to language-conditioned multitask control. Action Chunking with Transformers (ACT) predicts action sequences to reduce compounding errors~\cite{zhao2023act},and Diffusion Policy models multimodal action distributions through iterative denoising~\cite{chi2023diffusion}. Building on these ideas, recent generalist policies scale language-conditioned control with large robot datasets and pretrained visual and language representations~\cite{brohan2022rt1,brohan2023rt2,octo2024,kim2024openvla,black2024pi0,pi05_2025,shukor2025smolvla}, achieving strong semantic transfer across tasks.

Despite this progress, where the policy looks remains largely hidden. In most architectures, task-relevant local visual evidence exists only inside a visual backbone or cross-attention layers, with no interface that exposes which image regions actually condition the generated actions. We argue that this implicitness has two practical consequences. First, the policy is difficult to inspect and debug. Second, it is fragile under visual change: LIBERO-plus shows that policies with high scores on standard benchmarks remain sensitive to camera pose, robot initialization, lighting, background texture, sensor noise, object layout, and language variation ~\cite{fei2025liberoplus}. We view these two symptoms as sharing one cause—the action model has no explicit channel that preserves local, task-aligned visual evidence.

We therefore ask: can an explicit, language-guided local feature map improve a multitask diffusion policy without changing its action objective? To answer this, we propose TLC-DiT, a plug-in extension of the Multitask DiT Policy~\cite{barreiros2026careful,lerobot_multitaskdit}. For every camera view, frozen DINOv2 patch features are modulated by the task instruction through FiLM and refined by a lightweight CoordConv CNN adapter, producing an explicit local feature map. This map is concatenated with the original global CLIP image condition, so that local evidence supplements the global context. The diffusion objective and action-generation procedure are left untouched.

Our contributions are threefold. First, we introduce an explicit local visual conditioning path that aligns dense DINOv2 features with the task instruction and injects them into a DiT action generator. Second, we design a small CNN adapter that adds two-dimensional coordinates and spatial smoothing while keeping the original diffusion training target unchanged. Third, we evaluate standard multitask performance, seven types of robustness perturbations, five module ablations, and two bimanual real-world tasks.

\section{Related Work}

\subsection{Imitation Learning and Generative Robot Policies}
ACT generates action chunks with a conditional variational transformer and is effective for precise bimanual manipulation~\cite{zhao2023act}. Diffusion Policy represents the action sequence as a conditional denoising process, which captures multimodal and high-dimensional behaviors~\cite{chi2023diffusion}. Behavior Transformer discretizes action modes and predicts continuous offsets~\cite{shafiullah2022bet}, and Implicit Behavioral Cloning adopts an energy-based formulation~\cite{florence2021ibc}. We build on the diffusion formulation for its multimodality, and use a transformer denoiser following the Diffusion Transformer design~\cite{lerobot_multitaskdit}, which serves as the action expert into which our local conditions are injected.

\begin{figure*}[t]
\centering
\includegraphics[width=0.98\textwidth]{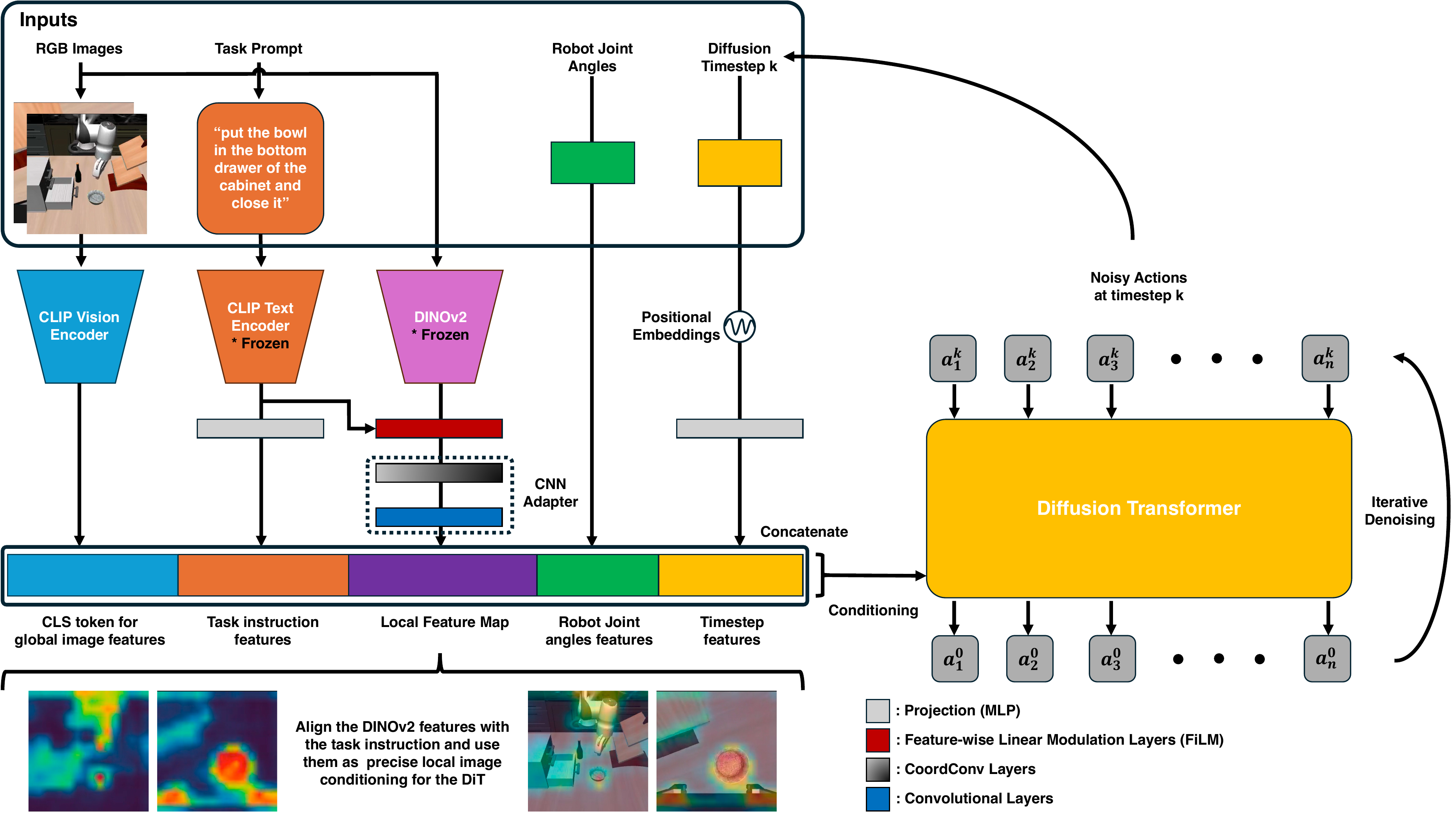}
\caption{Overview of \method. The original Multitask DiT global CLIP image feature, text feature, joint-state feature, and timestep feature are kept. The proposed path extracts frozen DINOv2 patch features, applies language-conditioned FiLM, and uses a CoordConv CNN adapter to form explicit local feature-map tokens. All conditions are concatenated for diffusion action generation. The values on the heat map range from low to high, with colors ranging from blue to red.}
\label{fig:architecture}
\end{figure*}

\subsection{Vision-Language-Action Models}
RT-1 and RT-2 scale language-conditioned control with diverse robot data and vision-language pretraining~\cite{brohan2022rt1,brohan2023rt2}. Octo provides an open generalist policy adaptable to new observation and action spaces~\cite{octo2024}, and OpenVLA combines pretrained language and dual visual representations~\cite{kim2024openvla}. Flow-based $\pi_0$ and its open-world extension $\pi_{0.5}$ focus on dexterous action generation and heterogeneous co-training~\cite{black2024pi0,pi05_2025}. SmolVLA reduces the model size and supports efficient asynchronous inference~\cite{shukor2025smolvla}. Across these works, improvements mainly concern semantic understanding, scaling, or action decoding, the visual evidence that conditions each action remains implicit in the backbone. TLC-DiT is complementary to this line: it can, in principle, be attached to any VLA-family policy to expose a task-related local visual signal as an independent condition.

\subsection{Local Visual Representation and Conditioning}
Explicit spatial structure has long been studied for control and grounding. Spatial autoencoders learn compact task-related keypoints for visuomotor control ~\cite{finn2016spatial}. CLIPort separates semantic ``what'' features from spatial ``where'' features~\cite{shridhar2021cliport}. In vision-language grounding, MDETR and Grounding DINO localize text-related objects via multimodal detection ~\cite{kamath2021mdetr,liu2024groundingdino}. TLC-DiT shares the motivation of explicit grounding but differs in two ways: it conditions a diffusion action generator directly on dense feature maps rather than on detected boxes or learned keypoints, and it requires no grounding supervision beyond the task instruction. Concretely, it combines CLIP for language semantics~\cite{radford2021clip}, DINOv2 for dense visual features~\cite{oquab2024dinov2}, FiLM for feature-wise language modulation~\cite{perez2018film}, and CoordConv for explicit position channels~\cite{liu2018coordconv}. The action backbone follows the Diffusion Transformer idea~\cite{peebles2023dit}.

\section{Method}
\subsection{Problem Formulation and Multitask DiT}
At time $t$, the policy receives multi-view RGB observations $\mathcal{I}_t=\{I_t^v\}_{v=1}^{V}$, robot joint state $s_t$, and a task instruction $l$. The target is an action chunk $A_t=[a_t,\ldots,a_{t+H-1}]\in\mathbb{R}^{H\times d_a}$. Multitask DiT uses a CLIP vision encoder for global image features, a frozen CLIP text encoder with learnable MLP for the task instruction, and a diffusion timestep embedding. These conditions are concatenated and provided to a transformer denoiser.

For diffusion step $k$, Gaussian noise $\epsilon\sim\mathcal{N}(0,I)$ is added to the clean action chunk:
\begin{equation}
A_t^k=\sqrt{\bar{\alpha}_k}A_t+\sqrt{1-\bar{\alpha}_k}\epsilon,
\label{eq:forward}
\end{equation}
where $\bar{\alpha}_k$ is defined by the noise schedule. The DiT predicts the added noise from the noisy actions and the observation condition $c_t$:
\begin{equation}
\hat{\epsilon}=\epsilon_{\theta}(A_t^k,k,c_t).
\end{equation}
The training objective is
\begin{equation}
\mathcal{L}_{\mathrm{diff}}=\mathbb{E}_{A_t,k,\epsilon}
\left[\left\|\epsilon-\epsilon_{\theta}(A_t^k,k,c_t)\right\|_2^2\right].
\label{eq:loss}
\end{equation}
At inference, an action chunk is iteratively denoised, and only the first configured action steps are executed before a new observation is used.

\subsection{Task-Aligned Local Feature Map}
Figure~\ref{fig:architecture} shows the proposed architecture. For camera $v$, a frozen DINOv2 (ViT-S/14) encoder produces a patch feature map
$F_t^v\in\mathbb{R}^{C\times h\times w}$. The language instruction is encoded by the frozen CLIP text encoder as $e_l$. A small projection predicts channel-wise scale and bias:
\begin{equation}
[\gamma_l,\beta_l]=\mathrm{MLP}_{\mathrm{film}}(e_l),
\end{equation}
followed by FiLM modulation
\begin{equation}
\widetilde{F}_t^v=\gamma_l\odot F_t^v+\beta_l.
\label{eq:film}
\end{equation}
The same task embedding is used for all camera views, while the visual features are view specific.

ViT patch features are spatially coarse and can appear discontinuous after reshaping. We therefore use a lightweight CNN adapter. Normalized horizontal and vertical coordinate maps $X,Y\in[-1,1]^{h\times w}$ are concatenated with the modulated features. The adapter is
\begin{equation}
M_t^v=\mathrm{Conv}_{1\times1}\!\left(
\mathrm{ReLU}\left(\mathrm{BN}\left(
\mathrm{Conv}_{3\times3}([\widetilde{F}_t^v;X;Y])
\right)\right)\right).
\label{eq:adapter}
\end{equation}
CoordConv gives direct two-dimensional position information, and the $3\times3$ convolution smooths neighboring patches. The output $M_t^v$ is flattened and linearly projected to local condition tokens $z_{\mathrm{loc}}^v$. The complete condition is
\begin{equation}
 c_t=[z_{\mathrm{img}}^1,\ldots,z_{\mathrm{img}}^V;
 e_l;z_{\mathrm{loc}}^1,\ldots,z_{\mathrm{loc}}^V;z_s;z_k],
\end{equation}
where $z_{\mathrm{img}}^v$ is the original CLIP vision encoder's CLS token for global image features, $z_s$ is the robot joint angles, and $z_k$ is the timestep feature. Thus, global context and explicit local evidence are jointly available to the DiT.

\subsection{Training}
We keep the diffusion objective in~\eqref{eq:loss} and the original Multitask DiT action scheduler. The CLIP text encoder and DINOv2 encoder are frozen. The CLIP vision encoder is fine-tuned with a smaller learning-rate (0.1) multiplier than the policy head. FiLM, the CNN adapter, projection layers, and the DiT are trained end to end. No extra local-map annotation is required. The baseline Multitask DiT has about 0.25B parameters, and \method{} has about 0.29B parameters. As shown in Table~\ref{tab:dit_configuration}, this configuration is shared by
the Multitask DiT, TLC-DiT, and all ablation models to
ensure a fair comparison. 

\begin{table}[h]
\centering
\caption{Diffusion Transformer configuration.}
\label{tab:dit_configuration}
\setlength{\tabcolsep}{6pt}
\renewcommand{\arraystretch}{1.05}
\begin{tabular}{lc}
\toprule
\textbf{Component} & \textbf{Configuration} \\
\midrule
Transformer dimension  & 512 \\
Transformer blocks     & 6 \\
Attention heads        & 8 \\
Dropout                & 0.1 \\
Timestep embedding     & 256 \\
Position encoding      & RoPE, base $10^{4}$ \\
\bottomrule
\end{tabular}
\end{table}

\section{Experiments}
\subsection{Simulation Benchmarks}
We evaluate on LIBERO~\cite{liu2023libero} and LIBERO-plus~\cite{fei2025liberoplus} using the LeRobot training framework~\cite{lerobot}. LIBERO contains four suites: Object, Spatial, Goal, and Long-10. Each suite consists of 10 tasks, for a total of 40 tasks. We compare SmolVLA (0.45B), Multitask DiT (0.25B), \method{} (0.29B), and five ablations. All simulation models use 2 observation steps, a horizon of 16, and 8 executed action steps. There are two input images, both with sizes of 224×224×3. They are trained for 100k gradient steps on the same LIBERO data until the training loss becomes stable. We trained each model with three random seeds, and selected the weight with middle-tier performance for testing. For Multitask DiT and all ablations, non-ablated components and settings are kept the same. 

LIBERO-plus evaluates seven perturbation groups: camera viewpoint, robot initialization, language, light, background texture, sensor noise, and object layout. A total of 10,030 tasks are included. All models were trained using only the LIBERO dataset and then tested directly on the LIBERO-plus evaluation environment. We report the success rate averaged inside each perturbation group and the total average.

\subsection{Real-World Bimanual Tasks}
\begin{figure}[h]
\centering
\includegraphics[width=0.48\textwidth]{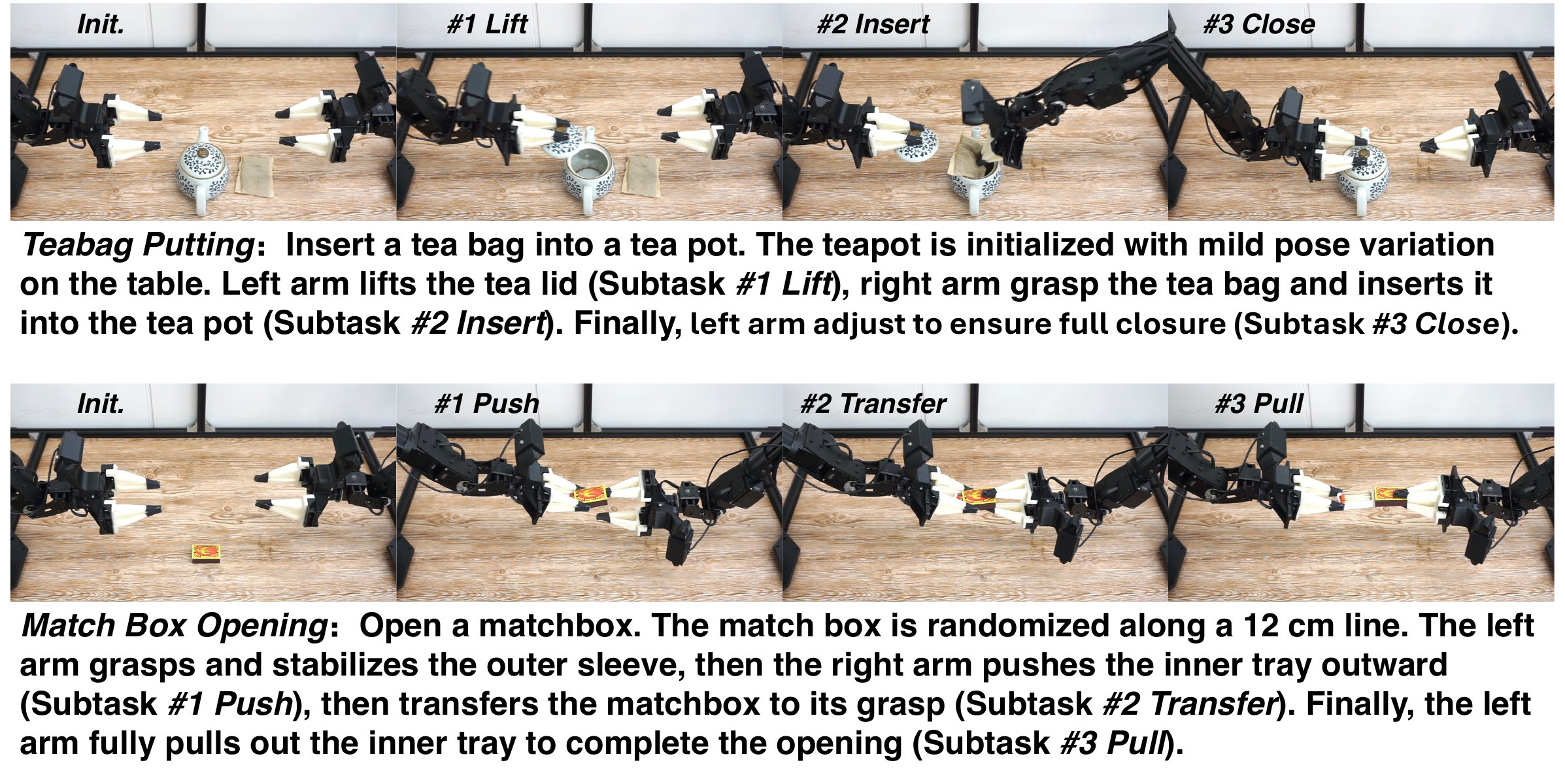}
\caption{Real-world tasks and evaluation milestones. Top: Teabag Putting (Lift, Insert, Close). Bottom: Match Box Opening (Push, Transfer, Pull).}
\label{fig:real_tasks}
\end{figure}

We use an ALOHA bimanual platform and collect four RGB camera (top, front, left arm, and right arm) streams at 640x480x3 together with 14 joint angles at 50~Hz. Figure~\ref{fig:real_tasks} shows the two tasks, \emph{Teabag Putting} and  \emph{Match Box Opening}. For each real-world task, we collect 50 human teleoperated demonstrations. Each demonstration is recorded for 20 seconds at randomized initial states along a 12 cm line. All policies use 1 observation step, an action horizon of 100, and 50 executed action steps. Each model follows its original image preprocessing and is trained for 100k steps until convergence. We compare ACT, SmolVLA, Multitask DiT, and \method, which have comparable parameters. Each task is evaluated for 100 trials per model. We report cumulative subtask success, so the final column is the full-task completion rate.

\begin{figure*}[t]
\centering
\includegraphics[width=0.8\textwidth]{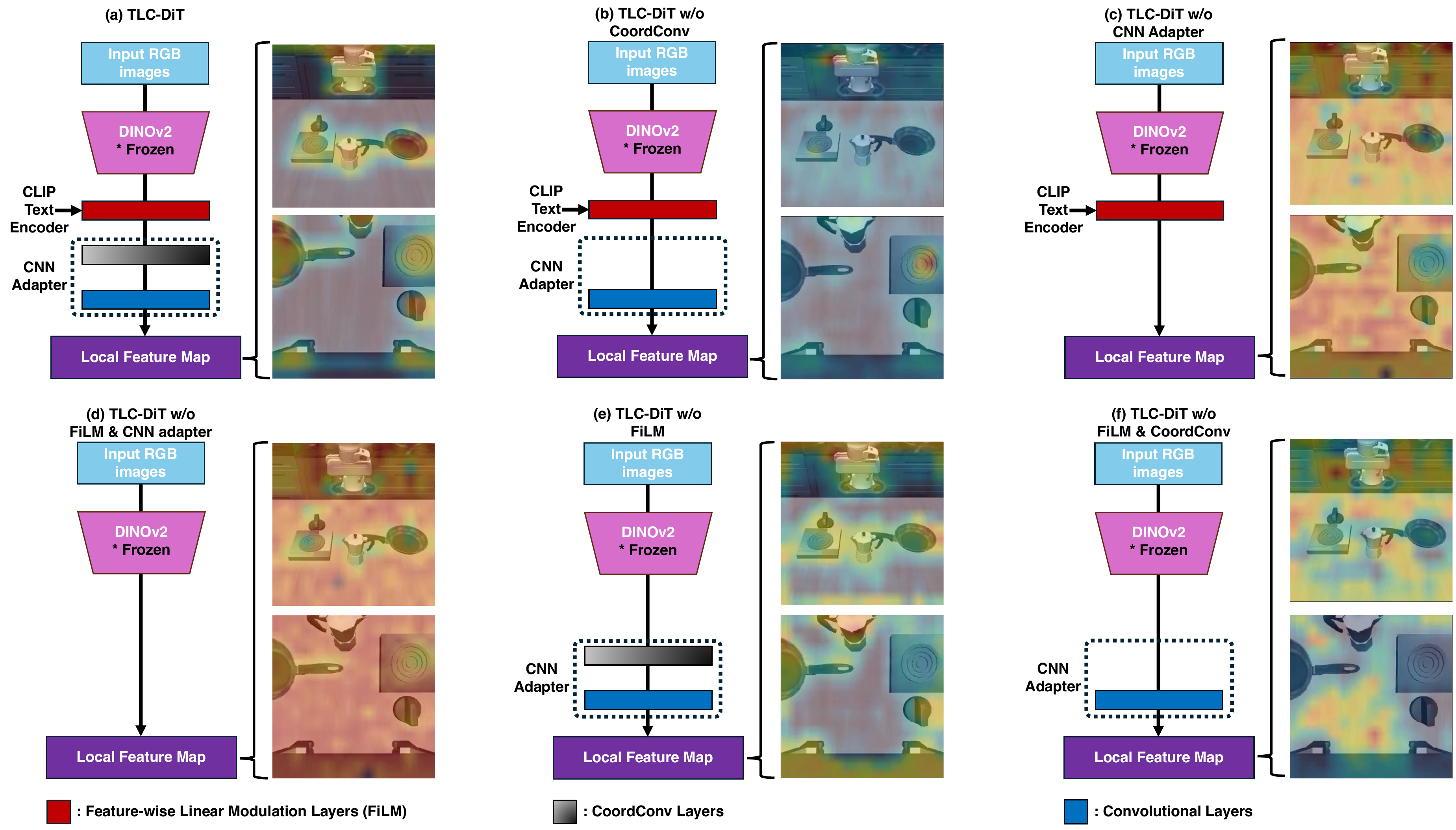}
\caption{Local feature map comparison for \method{} and five ablations. Each example shows two camera views. FiLM provides task-dependent modulation, while the CNN adapter and CoordConv produce cleaner and more spatially organized maps. The values on the heat map range from low to high, with colors ranging from blue to red.
}
\label{fig:ablation_maps}
\end{figure*}

\section{Results and Discussion}
\subsection{LIBERO Multitask Performance}
Table~\ref{tab:libero} reports success rates on the four LIBERO suites. Each suite is tested 100 times, for a total of 400 tests. \method{} reaches 93.5\% average success, improving Multitask DiT by 7.0 percentage points and SmolVLA by 14.25 points. The largest improvement over Multitask DiT is on Long-10, from 75\% to 87\%. This result suggests that the explicit local path is useful when a task needs several stages and object relations. Improvements on Object and Goal are both four points, while Spatial improves by eight points.

The full model gives the best overall average among the ablations. Removing the CNN adapter reduces the average to 91.5\%, and removing FiLM or CoordConv gives 92.0\%. Some ablations are stronger on an individual suite. For example, the model without FiLM and CoordConv reaches 88\% on Long-10 but drops to 89\% on Goal. Therefore, the modules do not improve every suite independently, but their combination gives the most balanced result. 

\begin{table}[t]
\caption{Success Rate (\%) on LIBERO}
\label{tab:libero}
\centering
\setlength{\tabcolsep}{5.0pt}
\begin{tabular}{lccccc}
\toprule
Model & Object & Spatial & Goal & Long-10 & Avg. \\
\midrule
SmolVLA (0.45B) & 89 & 79 & 88 & 61 & 79.25 \\
Multitask DiT (0.25B) & 94 & 86 & 91 & 75 & 86.50 \\
\textbf{\method{} (0.29B)} & \textbf{98} & 94 & \textbf{95} & 87 & \textbf{93.50} \\
\midrule
(b) w/o CoordConv & 95 & 94 & 93 & 86 & 92.00 \\
(c) w/o CNN adapter & 95 & 95 & 90 & 86 & 91.50 \\
(d) w/o FiLM \& CNN & 97 & 93 & \textbf{95} & 85 & 92.50 \\
(e) w/o FiLM & 96 & 94 & 92 & 86 & 92.00 \\
(f) w/o FiLM \& Coord. & 97 & \textbf{96} & 89 & \textbf{88} & 92.50 \\
\bottomrule
\end{tabular}
\end{table}

\subsection{Effect of Local Feature Map Components}
Figure~\ref{fig:ablation_maps} compares local maps for the full proposed model and ablations on the same LIBERO scene. The complete model produces concentrated responses around task-related objects in both views. Without the CNN adapter, the response remains broad and patch-like. Without FiLM, several objects receive similar activation because the map lacks direct task-language modulation. Removing CoordConv changes the spatial distribution and makes boundaries less stable. These visual differences are consistent with the average performance drop in Table~\ref{tab:libero}.

\subsection{Robustness on LIBERO-plus}

\begin{figure}[h]
\centering
\includegraphics[width=0.48\textwidth]{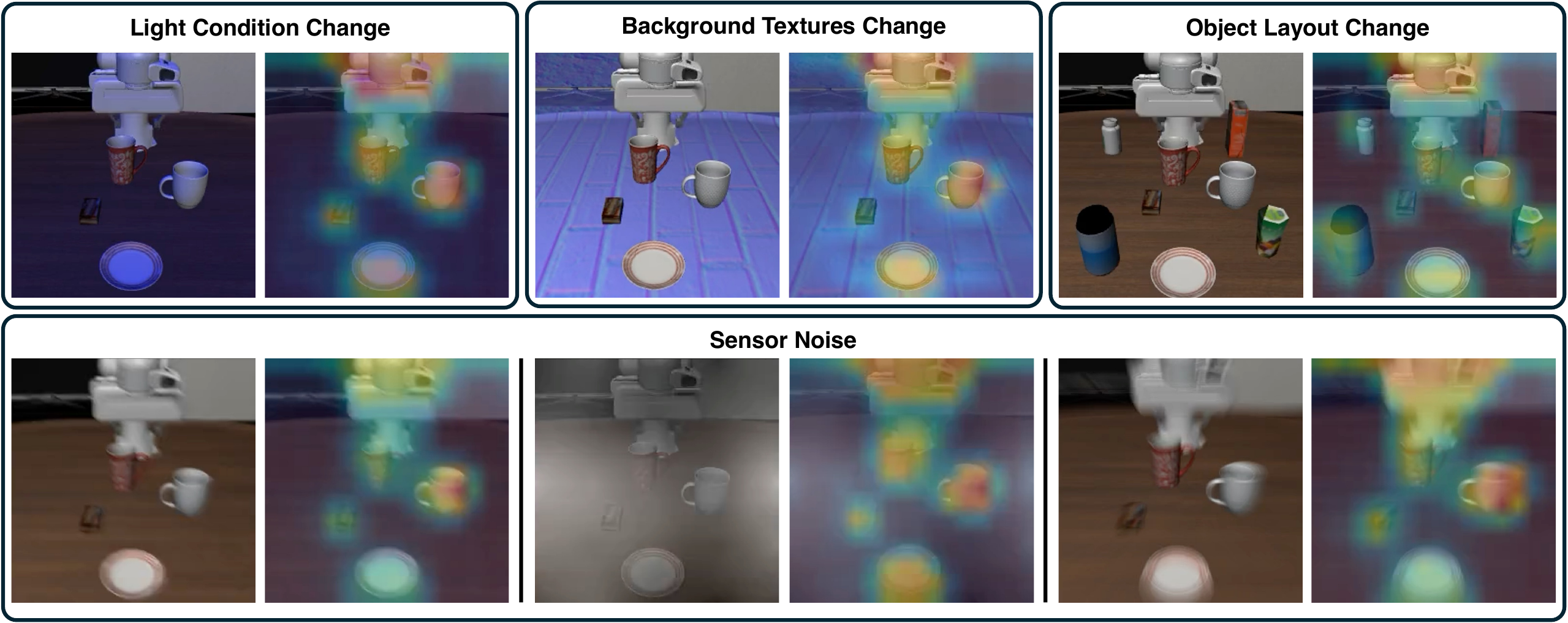}
\caption{Representative \method{} local feature maps under LIBERO-plus perturbations. From left to right and top to bottom: light condition, background texture, object layout, and three sensor-noise examples.}
\label{fig:liberoplus_maps}
\end{figure}

Figure~\ref{fig:liberoplus_maps} gives representative local feature maps under light, background, layout, and sensor-noise changes. The activated regions remain near the robot gripper and task objects even when appearance changes. Under strong blur, the map becomes more diffuse, which matches the remaining performance gap under sensor noise.

\begin{table*}[t]
\caption{Success Rate (\%) under LIBERO-plus Perturbations}
\label{tab:liberoplus}
\centering
\setlength{\tabcolsep}{8pt}
\begin{tabular}{lcccccccc}
\toprule
Model & Camera & Robot & Language & Light & Background & Noise & Layout & Total \\
\midrule
Multitask DiT & 42.71 & 34.65 & 72.54 & 68.13 & 52.88 & 46.41 & 65.44 & 54.07 \\
\textbf{\method} & \textbf{48.78} & 37.35 & 73.52 & 68.65 & \textbf{58.55} & \textbf{50.97} & 67.02 & \textbf{57.24} \\
\midrule
(b) w/o CoordConv & 47.15 & 37.03 & 72.74 & 69.09 & 52.88 & 46.97 & 66.62 & 55.55 \\
(c) w/o CNN adapter & 45.28 & \textbf{38.06} & 71.57 & 67.25 & 54.83 & 49.28 & 66.69 & 55.61 \\
(d) w/o FiLM \& CNN & 43.03 & 35.03 & \textbf{74.04} & \textbf{72.07} & 57.90 & 39.29 & 65.18 & 54.22 \\
(e) w/o FiLM & 44.84 & 34.65 & 72.54 & 68.21 & 51.86 & 42.54 & \textbf{70.95} & 54.53 \\
(f) w/o FiLM \& Coord. & 43.84 & 35.03 & 70.27 & 67.08 & 55.30 & 45.53 & 64.39 & 53.80 \\
\bottomrule
\end{tabular}
\end{table*}

\begin{figure*}[t]
\centering
\includegraphics[width=0.85\textwidth]{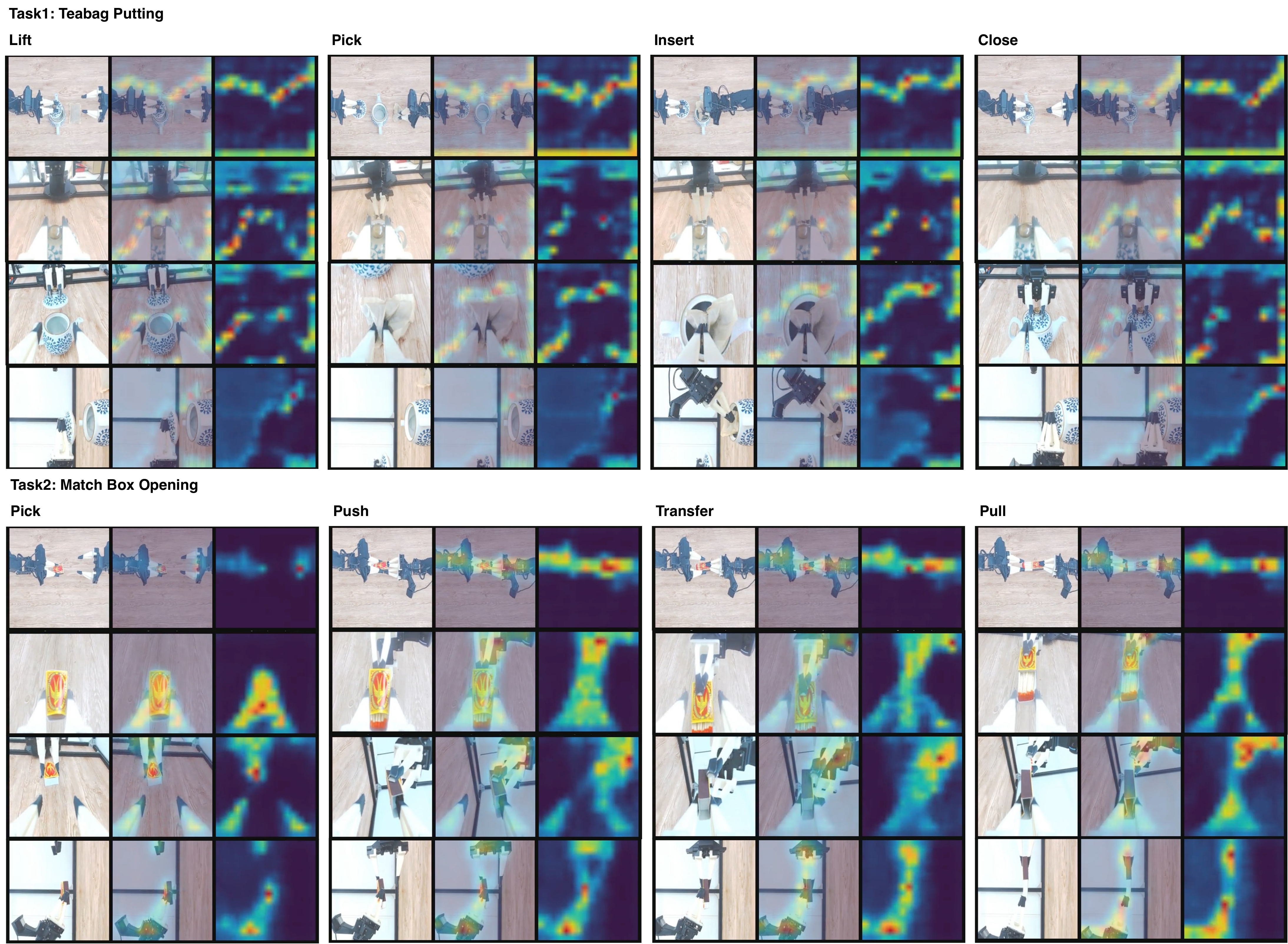}
\caption{Multi-view local feature maps during real-world execution. For each temporal phase and camera view, the columns show the RGB image, feature overlay, and local map. The values on the heat map range from low to high, with colors ranging from blue to red.}
\label{fig:real_maps}
\end{figure*}

Table~\ref{tab:liberoplus} reports results under the seven LIBERO-plus perturbation groups. \method{} improves the total rate from 54.07\% to 57.24\%. It is higher than Multitask DiT in all seven groups. The main gains are camera viewpoint (42.71\% to 48.78\%), background texture (52.88\% to 58.55\%), and sensor noise (46.41\% to 50.97\%). The gains for language and light changes are smaller. This is reasonable because the added path mainly changes local visual conditioning rather than the language encoder or action objective.

The ablation order is clearer under perturbations than on standard LIBERO. Removing the CNN adapter gives 55.61\%, removing CoordConv gives 55.55\%, and removing FiLM gives 54.53\%. Removing FiLM together with CoordConv gives 53.80\%, below the original Multitask DiT. This indicates that dense DINOv2 features alone are not sufficient, task modulation and spatial adaptation are both needed. At the same time, the final 57.24\% rate shows that robustness is still far from solved, especially for robot initialization and camera changes.

\subsection{Real-World Results}
Table~\ref{tab:real} shows cumulative stage success in 100 trials. On Teabag Putting, \method{} keeps 100\% Lift success and increases Insert from 50\% to 92\% and full Close completion from 44\% to 89\% compared with Multitask DiT. It also outperforms ACT and SmolVLA at every scored stage. This task requires selecting a small deformable tea bag, locating the teapot opening, and coordinating two arms. The explicit local condition is helpful for these object-level relations.

On Match Box Opening, \method{} is comparable with Multitask DiT. It is slightly lower on Push and Transfer (76\%/69\% versus 80\%/70\%) and slightly higher on full Pull completion (55\% versus 52\%). Both diffusion-transformer policies are better than ACT and SmolVLA on final completion. The small difference suggests that this task is limited not only by visual grounding but also by contact dynamics and precise force interaction, which are not directly addressed by the proposed visual module.

Figure~\ref{fig:real_maps} shows four representative temporal snapshots from each task. Each camera row contains the RGB input, its overlay, and the local feature map. The maps shift with the current interaction: from the arms and lid to the tea bag and teapot opening, and from the matchbox body to the tray and pulling region. The snapshots include an initial approach or picking phase in addition to the three scored milestones of Table~\ref{tab:real}. The local feature maps show a tendency similar to that observed in simulation. In Match Box Opening, TLC-DiT produces stronger responses around the task-related interaction regions and can identify the matchbox even though it occupies only a small part of the image. In Teabag Putting, the local maps emphasize the boundaries of the tea bag, teapot opening, and lid. These boundary-sensitive features may help represent the spatial relations required for insertion.

\begin{table}[t]
\caption{Cumulative Success Rate (\%) on Real-World Tasks}
\label{tab:real}
\centering
\footnotesize
\setlength{\tabcolsep}{6pt}
\begin{tabular}{lccc ccc}
\toprule
& \multicolumn{3}{c}{Teabag Putting} & \multicolumn{3}{c}{Match Box Opening} \\
\cmidrule(lr){2-4}\cmidrule(lr){5-7}
Model & Lift & Insert & Close & Push & Transfer & Pull \\
\midrule
ACT & 65 & 46 & 21 & 76 & 47 & 38 \\
SmolVLA & 74 & 73 & 56 & 12 & 0 & 0 \\
Multitask DiT & \textbf{100} & 50 & 44 & \textbf{80} & \textbf{70} & 52 \\
\textbf{\method} & \textbf{100} & \textbf{92} & \textbf{89} & 76 & 69 & \textbf{55} \\
\bottomrule
\end{tabular}
\end{table}

\section{Conclusion}
We presented \method, an explicit language-guided local visual conditioning path for Multitask DiT. The method combines frozen DINOv2 patch features, CLIP-conditioned FiLM, and a CoordConv CNN adapter, and injects local feature maps without changing the diffusion action objective. It improves average LIBERO performance and gives consistent gains across all LIBERO-plus perturbation groups. For ALOHA real-world tasks, it strongly improves Teabag Putting and maintains comparable Match Box Opening performance. The local maps also provide a direct way to inspect which image regions are supplied to the action model. Current limitations include modest gains on contact-dominated tasks, remaining sensitivity to camera and robot initialization. Future work will combine explicit local grounding with long-horizon planning, world models, and execution verification.

\addtolength{\textheight}{-1cm}   




\section*{ACKNOWLEDGMENT}

This work was supported in part by JST BOOST, Japan, under Grant JPMJBS2429; in part by the JST Moonshot Research and Development Program, Japan, under Grant JPMJMS2031; and in part by the Research Institute for Science and Engineering, Waseda University.

\bibliographystyle{IEEEtran}
\bibliography{refs}

\end{document}